**Empirical Translation Process Research: Past and Possible Future Perspectives**

Michael Carl, Kent State University

**Abstract**

Over the past four decades, efforts have been made to develop and evaluate models for Empirical Translation Process Research (TPR), yet a comprehensive framework remains elusive. This article traces the evolution of empirical TPR within the CRITT TPR-DB tradition and proposes the Free Energy Principle (FEP) and Active Inference (AIF) as a framework for modeling deeply embedded translation processes. It introduces novel approaches for quantifying fundamental concepts of Relevance Theory (relevance, s-mode, i-mode), and establishes their relation to the Monitor Model, framing relevance maximization as a special case of minimizing free energy. FEP/AIF provides a mathematically rigorous foundation that enables modeling of deep temporal architectures in which embedded translation processes unfold on different timelines. This framework opens up exciting prospects for future research in predictive TPR, likely to enrich our comprehension of human translation processes, and making valuable contributions to the wider realm of translation studies.

**Keywords**: translation process research; relevance theory; monitor model; free energy principle; active inference

## 1 Introduction and historical background

In the 1970s, Translation Studies emerged as an empirical, scientific (descriptive) discipline. By the mid-1980s a branch in of this field now often referred to as Cognitive Translation Studies (CTS, or more recently CTIS, Cognitive Translation and Interpretation Studies) started to investigate and model how the translators' minds work –how translators create meaning, how they arrive at strategies and translation choices, how translation competence develops, how cultural and linguistic factors impact translated text, etc. (see, e.g., Risku 2012). Studies in this line of research "refer to and expand" (Risku 2012, 675) models of the mind as developed in Cognitive Science, to explain translators' behavior and translation processes.

While the first attempts to study translation as a cognitive activity date back to the 1960s and 1970s (e.g., Albir 2015, Muñoz 2017), Translation Process Research (TPR) is often said to begin in the 1980s with the analysis of thinking aloud protocols (TAP) and to investigate "What happens in the minds of translators" (Krings 1986; 2001; see also Königs 1987) and to assess "by what observable and presumed mental processes do translators arrive at their translations?" (Jakobsen 2017, 21). Empirical TPR thereby uses a range of technologies to record and analyze behavioral data.

Since the 1980s and early 1990s, TPR has evolved in several phases with the increasing availability and usage of new sensor and tracking technologies, suitable for recording and analyzing the translation process. A new era in TPR was introduced in 1996 with the development of a special-purpose software Translog (Jakobsen and Schou 1999) and the TRAP (Translation Process) project, in which

> researchers from different language departments at the Copenhagen Business School (CBS) launched a translation project with the aim of promoting research into the translation process … [as] it was felt that our understanding of the mental processes could be improved if the traditional qualitative approaches could be supplemented by quantitative data.
> Hansen 1999, 7

Successively, Muñoz (2010) suggested ten principles by which he defined a new framework for *cognitive translatology*, which would ground translation research, among other things, in empirical data. Muñoz made out an "urgent need to establish experimental paradigms" (169) in which "research must be firmly grounded in observable translation reality" (174). Coincidently, these needs were addressed around the same time at the Center for Research and Innovation in Translation and Translation Technology (CRITT) which started gathering behavioral data and making it available to the public as a database, the CRITT TPR-DB.

The CRITT TPR-DB (Carl et al. 2016) has been a major endeavor to gather and compile Translation Process Data (TPD) into a consistent and coherent format that can be used for advanced analysis of the process. The TPR-DB started within the EU Eye-to-IT project (2005-2009) and successively CASMACAT (2011-2014) in which several Ph.D. projects collected numerous datasets that were then integrated into an experimental database. The CRITT TPR-DB has thus evolved into an open-access framework with several possibilities for further extension into various directions, to accommodate diverse data-acquisition tools (e.g., Translog-II, Carl 2012, and CASMACAT; Alabau et al. 2014), and more recently Trados (Zou et al. 2022, Yamada at al. 2022), to prototype new features and explore different explanatory models of the translation process.

Initially, a query language had been conceived to provide several basic operators that could be composed into more complex functions to generate and extract features from the raw TPR-DB data (Carl and Jakobsen 2009). However, this quickly turned out to be impractical, due to the size of the data, the processing time, and the complexity of the operators. Rather than implementing a (meta) language to generate ad-hoc features on the fly, the operations were aggregated in a separate processing step to generate a set of summary tables. These summary tables are now an integral component of the TPR-DB; they list a large number of product and process features that are used for data analysis[1].

Jakobsen (2017) distinguishes three phases in the development of TPR: the TAP phase, a keylogging and eye-tracking phase, and more recently the integration and deployment of methods originating in data-analytics and data-sciences. Jakobsen says:

> TPR has been dominated methodologically, first by the use of introspective methods, primarily TAPs and retrospection, then by (micro-) behavioral methods, keylogging and eye-tracking, sometimes in combination with cued retrospection, and more recently by the application of computational methods in the analysis of very large amounts of process data.
> Jakobson 2017: 39

A large body of data and research findings have been produced in the past two decades. More than 5000 translation sessions and hundreds of hours of TPD have been recorded so far. A part of the data is publicly accessible to everyone, free of charge; the raw logging data is permanently stored in a public repository[2]. Users of the TPR-DB can obtain a personal account to organize their data in studies. Studies consists of one or more, sometimes hundreds of, (translation) sessions that contain the logged data. Each session comprises 11 summary tables with a total of more than 300 features describing properties of the translation process and the translation product (see Carl et al. 2016). The summary tables are under constant revision and extension, as new features are added, and summary tables re-generated. A browser interface has been put in place with direct access to the TPD and a Jupyter/python toolkit has been set up that allows for advanced data-analytics methods in empirical TPR.

---

[1] The features are described on the CRITT website: https://sites.google.com/site/centretranslationinnovation/tpr-db/features. The technological repository of data collection methods have dramatically increased in the last decade and includes, besides keylogging and eye-tracking, also EEG, fMRI, and fNIRS technologies. Such data is, however, not part of the TPR-DB.

Based on the TPR-DB framework numerous studies[3] have investigated the relation between the translation processes and the product (i.e., behavioral and linguistic data), personal and demographic properties of translators (expertise, experience, education, etc.), as well as translation goals (e.g., translation guidelines) and methods of the translation tasks (e.g., from-scratch translation, post-editing, etc.). These studies investigate, among other things, the role of expertise, and translation directionality (L1/L2 translation), ergonomic, linguistic, and emotional factors, as well as the usage of (external) resources - such as computer assisted translation and machine translation (MT). TPR focuses thereby primarily on process related issues, such as temporal aspects (e.g., translation duration), translation effort, and the distribution of (visual) attention as e.g., gathered by eye-tracking devices.

Several competing theoretical approaches have been deployed to explain the reported observations. Hvelplund (2011) draws on theories of working memory (Baddeley 1974; 2000) and of a central executive system (Baddeley 2007) to explain translation processes, while Sjørup (2013) and Schmaltz (2015) refer to Lakoff and Johnson (1980) to assess the cognitive effort in metaphor translation. Serbina (2015) and Heilmann (2020) deploy Cognitive and Systemic Functional Grammar for understanding and explaining translation processes, while Alves and

---

[2] Source forge SVN repository. Instructions are available on https://sites.google.com/site/centretranslationinnovation/tpr-db/public-studies

[3] For an overview see publications on https://sites.google.com/site/centretranslationinnovation/tpr-db-publications

Vale (2009) develop an empirical interpretation of translation units (TUs) to ground Relevance Theory (RT, Gutt 1991; 2000) in empirical behavioral data. Schaeffer and Carl (2013/2015) suggest a *Monitor Model* that builds on findings of Tirkkonen-Condit (2005) and aspects of bilingualism (Halverson 2003; 2010). A further development of the Monitor Model (Carl and Schaeffer 2019) joins it with insights from RT.

In the meantime, post-cognitivist approaches have emerged within the field of CTS that account for the extended and enacted nature of translation (Risku 2012; Risku and Rogl 2020; Carl 2021; Muñoz 2021), while Schaeffer et al. (2020, 3939) announce the "predictive turn in translation studies" in which machine learning approaches would be used for modelling human translation processes and for predicting "when and why a translator is having trouble carrying out the [translation] task" (3940).

In this article I argue that the Free Energy Principle (FEP, Friston 2009; 2010) and Active Inference (AIF, Parr et al. 2022) constitute a suited framework to account for these new demands: FEP and AIF are formulated within a mathematical rigorous framework (i.e., Bayesian reasoning) in which (artificial) agents are modelled as Partially Observable Markov Decision Processes (POMDP) that can learn from experiences, and it has been analyzed within an enactivist framework (Bruineberg et al. 2018; Kirchhoff and Kiverstein 2021; Carl 2023).

FEP and AIF constitute a general unifying perspective of how biological organisms maintain a balance between their internal states and the external environment, necessary to ensure their survival in an ever-changing environment. I argue that AIF may also be a promising pathway to advance (predictive) TPR. Several TPR models imply a "deep temporal architecture" (Parr et al. 2023), in which concurrent processes in distinct temporal strata complement and interact with each other. The Monitor Model (Schaeffer and Carl 2013/2015), for instance, assumes two processes, i.e., automatized translation routines are at the basis of human translation production until "interrupted by a monitor that alerts about a problem in the outcome. The monitor's function is to trigger off conscious decision-making to solve the problem"[4] (Tirkkonen-

---

[4] Her paper does not commit what exactly is meant by "decision-making" and "problem".

Condit 2005, 11). RT defines translation as interpretive language use and suggests two distinct translation modes, a "stimulus mode" and an "interpretive mode" that, I argue –similar to the automatized and monitoring processes of the Monitor Model– unfold in different timelines, while Muñoz and Apfelthaler (2022) suggest a "task segment framework" in which assumed intentional and unintentional processes generate pauses of different length.

In section 2, I extend the Monitor Model with elements of RT. I argue that Gutt's (2004, 2005) distinction between *stimulus mode* and *interpretive mode* roughly fits the dichotomy between translation automaton and monitor processes as suggested in the Monitor Model, in which monitoring processes control the interlingual resemblance of the source and the target.

Section 3 elaborates a novel operationalization of relevance. Relevance is a central concept in RT, defined as a function of effort and effect (Sperber and Wilson 1995, 132). Communication, in their view, is geared towards the maximization of relevance, but, as I will show, the concept has been deployed in various ways in translation, where s-mode and i-mode translations imply different presumptions of relevance. While RT stipulates that relevance cannot be quantified, I introduce a new notion, the *field of relevance*, opening the possibility for quantifying and determining optimal relevance. In section 4 I argue that relevance is inversely proportional to *free energy*, which is, according to FEP, a quantity that living agents (such as translators) need to minimize. In conclusion, I argue that the FEP and AIF are suited to accommodate those notions (and many more) in a rigorous mathematical framework, which may be suited to advance predictive TPR in the next decade or so.

## 2 The Monitor Model

The Monitor Model, as proposed by Schaeffer and Carl (2013/2015), stipulates that automatized priming routines are the basis of human translation processes. Priming processes activate and integrate entrenched translation patterns which map ST expressions into (default) TL equivalents. In addition, on another timeline, the Monitor Model assumes that higher order monitoring strategies provide translators with criteria to decide whether the produced translations actually correspond to the translation aims, guidelines or goals. While priming is quick and associated with low levels of effort, monitoring processes may take up a substantial amount of time and higher levels of translation effort. For instance, monitoring processes seem to disintegrate loops

To be published in Translation, Cognition and Behavior: "Translation and cognition in the 21st century: Goals met, goals ahead"

of concurrent perception-action into successive reading and typing activities (cf. Carl and Dragsted 2012).

Loops of (ST) reading and (TT) typing in translation production have been referred to as TUs and constitute the empirical basis for some of the modelling in TPR (e.g., Alves and Vale 2009; Carl and Kay 2011). TUs are thought to be basic units by which translations are produced. Malmkjær (1998), for instance, defines process oriented TUs as a "stretch of the source text that the translator keeps in mind at any one time, in order to produce translation equivalents in the text he or she is creating" (286). While Malmkjær thinks of TUs as basically mental constructs, TUs have also been conceptualized as a combination of both, mental events and physical (observable) behavioral acts that exhibit properties of automatized and/or monitoring processes.

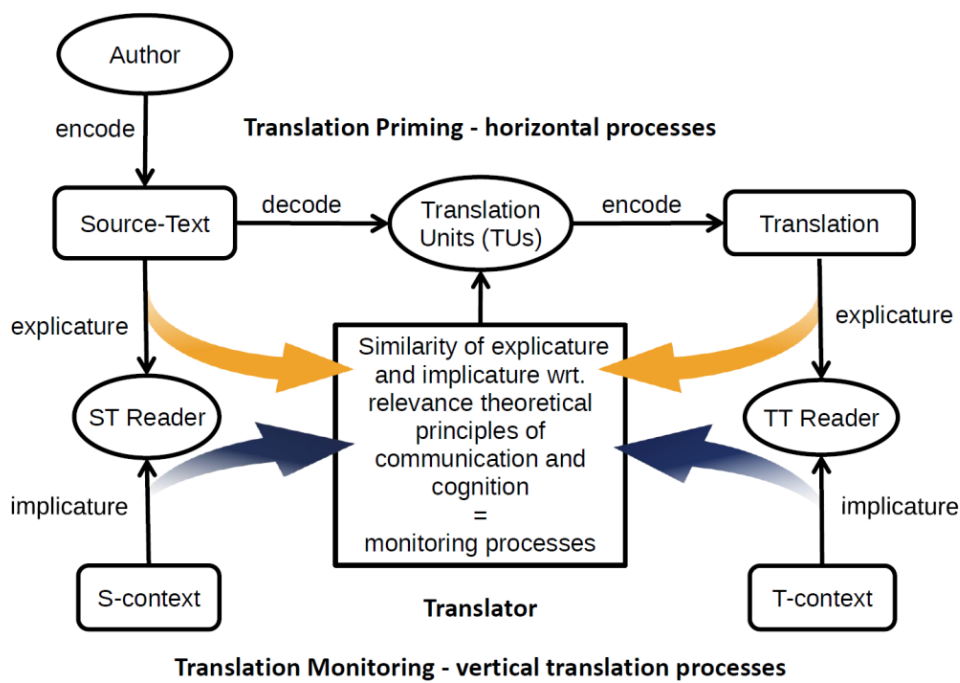

Figure 1: The Monitor Model, adapted from Carl and Schaeffer (2019)

An adaptation of the Monitor Model is plotted in Figure 1 (adapted from Carl and Schaeffer 2019). The model stipulates that the translation process unfolds in terms of TUs, while the similarity of the input and output is controlled on a higher-level by monitoring processes which compare the source and the target, for instance, as to whether translation guidelines are met. The model draws on Relevance Theory (RT) to address the importance of monitoring functions that ensure translation goals (e.g., translation quality, style, etc. as specified in translation guidelines) are achieved despite possibly diverging source and target contexts. It depicts the interaction of automatized horizontal and vertical monitoring processes.

According to RT, communication –as well as translation as a special form of interlingual communication– unfolds within the cognitive environment of an SL speaker and a TL hearer, mediated by the translator, where "the relevance-theoretic account brings out the **crucial role which context plays in translation**" (Gutt 2004, 3 emp. original). In order to make full use of the principle of relevance (see section 3), translators need to "meta-represent" the cognitive environment of the SL speaker, the TL audience and the context of the translation. Different translation strategies can be used depending on how much the cognitive environment in the SL and the TL overlap, where "a clear understanding of the influence of context can equip translators ... [with] the possibilities and limitations of translation as a mode of interlingual communication" (Gutt 2004, 3). Gutt (2005) describes several scenarios in which the SL author, the translator and the TL receptor share the same or a different cognitive environment to different degrees. Gutt (2004; 2005) introduces a distinction between a *stimulus* mode (s-mode) and an *interpretive* mode (i-mode) of translation production that result in different kinds of inter-lingual resemblance, where "the resemblance does not have to be between the intended interpretations [i-mode] but can also lie in the sharing of linguistic properties [s-mode]" (Gutt 2004, 4).

Provided the cognitive environment of the SL and the TL audience overlaps to a large extent, replication of stimulus characteristics into the TL (s-mode translations) may be sufficient for the informed target audience to re-construct the interpretive resemblance with the source: "it seems not unreasonable to consider an expression of language B a token of an expression of language A to the extent that they have properties in common" (Gutt 2005, 40). An informed receptor can recover the intended meaning from the traces of the stimulus when the translator informs the audience merely of the evidence, rather than the meaning. The i-mode, in contrast, may help translators bridge communication barriers, if the cognitive environment and/or the context between the SL and the TL audience is vastly different.

Gutt is not explicit about the interaction between the s-mode and the i-mode. The Monitor Model (Figure 1) models two processes as horizontal (priming) and vertical (monitoring) processes respectively which, I will argue, have properties of the s-mode and i-mode. In this view, s-mode translation is the basis of translational activity, on top of which i-mode translation may (or may not) adjust any potential communication gaps that may be left unaddressed by the s-mode.

## 3    Measuring Relevance

RT posits that a message is relevant if it connects with the background of the cognitive environment and the available contextual information to answer a question or a doubt, to confirm/reject a suspicion, or correct an impression, etc. thus to yield conclusions that matter. The relevance of an input can be assessed as a function of effort and effect (Gutt 1989, 50):

1. if the contextual effects in the given context are large
2. if the effort required to process the input in this context is small

RT postulates a cognitive principle of relevance according to which human cognition tends to maximize relevance, and a communicative principle according to which an act of communication presumes optimum relevance (Sperber and Wilson 1995, 260). From these principles follows that expectations of relevance raised by an utterance are precise and predictable so as to guide the hearer toward the speaker's meaning. A speaker (or writer) will seek at formulating their utterances such that the first interpretation that a hearer generates conveys the intended meaning. A speaker will stop whenever s/he thinks the desired effect may be retrieved by the hearer, thereby minimizing her effort. A hearer (or reader), on the other hand, follows a path of least effort and stops whenever a worthwhile effect has been generated (Sperber and Wilson 1995; Gutt 1989; 1991; 2000). Thus, RT defines relevance as a function of effort and effect but has not developed measures for quantification. In the following section, I suggest conceiving relevance within a two-dimensional field of relevance, as shown in Figure 2. I will point out a number of factors and contexts that may play a role in the quantification of relevance. In section 4, I present a formal framework that, I believe, is suited to capture essential components of RT and the Monitor Model.

### 3.1    *The field of Relevance*

Figure 2 depicts several hypothetical paths of relevance, as a trade-off between effort and effect. The upper left rectangle indicates areas of high relevance that can be achieved by expending an acceptable amount of effort while creating worthwhile effects. The lower right rectangle amounts to irrelevant cases that require too much effort and provide unworthy effects.[5] The red line marks a (hypothetical) boundary between levels of relevance and amounts of effort and effect. The curvature of the green lines suggests that the relation between effort and effect might be non-linear, dotted lines indicate unsuccessful paths through the field of relevance. It suggests that effort increases in time, while effects relate to changes in system configurations (see Table 1 and

Table 2).

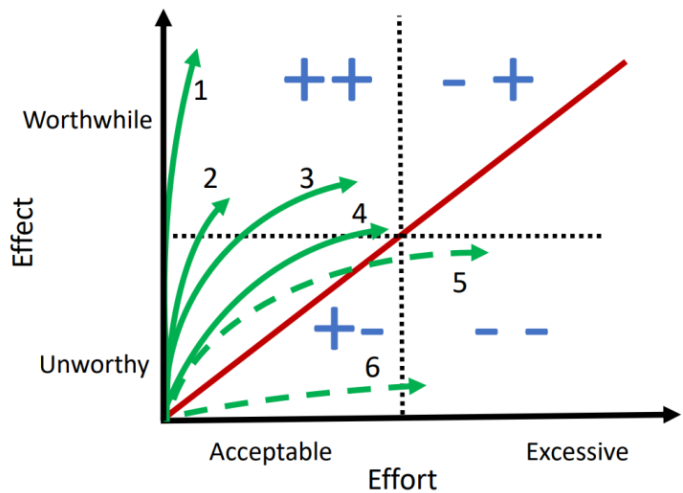

Figure 2: The field of Relevance as trade-off between effort and effect

While RT has originally been conceived as a theory of communication in a monolingual setting, the relevance criteria have also been applied to explain translation. In the translation context, it may be easy for a translator to generate the worthwhile intended effects of a text in another target language if the source speech/text is simple and/or both the source and the target audience share the same cognitive environment. In this case, a translator can make use of the s-mode if the target audience can recover the intended meaning from the transposition of the stimulus. Such instances may have a shape similar to relevance path **1** in Figure 2. However, a translation is likely to become increasingly difficult and time consuming to generate if, for instance, the utterance is ambiguous or unclear, when the context is unavailable, implicatures cannot be retrieved and/or the cognitive environment in the source and target are very different. Such challenging translations may exhibit a relevance path similar to path **4** in Figure 2. Memory constraints may play a role if utterances are too long or too complex, and/or the required inferences are not obvious or unachievable in the current situation.

---

[5] Note that the terminology, "worthwhile", "unworthy", "acceptable", "excessive" is taken from Gutt (and RT) while the formulation of the field of relevance is my own.

However, irrespective of the context, a hearer, reader, or translator will stop processing as soon as a satisfying interpretation is generated (or believed to the retrievable for the TL audience). Or otherwise, according to RT, processing efforts may be given up all together when a worthwhile effect cannot be reached within an acceptable amount of effort, as exemplified in relevance paths **5** and **6**.

Gile and Lei (2020) point out that a balance between effort and effect is central in translational action. They assume a strong correlation between effort and effect for low effort which then flattens out as higher amounts of effort are invested. "Beyond a certain point, further effort may not contribute much or may even become counterproductive, though where this point is can vary greatly." (265). In Gile and Lei, a distinction between low-intensity and high-intensity effort is made. Low-intensity effort occurs in activities such as terminological search, prolonged preparation of glossaries, and repeated revision. High-intensity effort has often been attributed to the exhaustion of non-automatic processes that draw on limited available resources during the translation process. High-intensity effort, they say, occurs basically only during interpretation.

### 3.2   *S-mode, I-mode and Relevance*

Gutt (2005) posits that the distribution of effort and effects are different for translators and audience under the s-mode and i-mode. Table 1 is adapted from (Gutt 2005); it summarizes parameters for these differences. S-mode translations are fast and easy *for the translator*, as, I suppose, they are largely based on priming processes (Carl 2023). I-mode translations, in contrast, can be expected to be more effortful for the translator, but are easier to comprehend for the audience.

According to Gutt, the main advantage of i-mode translations consists in the ease of comprehension for the TL audience, but it also bears risks as the translated message heavily depends on the interpretation, understanding, relevance judgements, and preferences of the translator, which may not always correspond to those of the SL speaker. The s-mode, in contrast, preserves a maximum amount of resemblance of the source and the translation; it implies less effort on the side of the translator but requires higher levels of awareness from the TL audience. In this view, provided the expected effects are sufficient, RT thus predicts that a translator will choose the s-mode whenever possible, as the expended translation effort is lower and the overall outcome is, thus, more relevant. It turns out that relevance from a translator's point of view is different as compared to the audience. They may not be symmetrical.

| | | *s-mode*–telling 'what was said' | *i-mode*–telling 'what was meant' |
|---|---|---|---|
| Effort / effect on audience | | *Effort for the Translator: Low* based on quick priming mechanisms | *Effort for the Translator: High* implies evaluation of context, reflective thought, search etc. |
| | Effort for audience | Potentially high (acquisition of context knowledge required; 'cautious optimism')* | Comparatively low (can simply use own context; 'naïve optimism')* |
| | Effects on audience | Contents not changed by translator to fit receptor context | Contents changed by translator to fit receptor context |
| | Meaning resemblance | High –independent of current cognitive environment of receptors | Variable –dependent on current cognitive environment of receptors |

* Sperber and Wilson (2002) define three degrees of metarepresentational ability, of which Gutt (2005) assumes two (1 and 2) to play a role for the audience of translations:
1) A Naively Optimist accepts the first relevant interpretation regardless of whether it could plausibly have been intended.
2) A Cautious Optimist is capable of dealing with mismatches of first-order false belief tasks, but unable to deal with deliberate deception.
3) A Sophisticated Understander has the capacity to deal simultaneously with mismatches and deception.

Option 3) can be excluded in a translation context, as it can be assumed that a translator will not attempt to deceive the target audience.

Table 1: differences between s-mode and i-mode, adapted from Gutt (2005).

### 3.3 *Shape of Relevance*

Effort, effect, and thus relevance may also be conceived of as multivariate continuous variables. For instance, an utterance may have an *effect* on one level, conveying important aspects of a message while lacking a worthwhile other effect on another level. An utterance may convey appropriate content (e.g., in terms of lexical items) but be syntactically or stylistically erroneous. It may express simple facts in a complicated manner or lack the required contextualization. Similarly, expenditure of effort may relate to retrieving low frequency words, ad-hoc metaphors, or unusual collocations, or it may be spent on analyzing long sentences or the disambiguation/contextualization of complex relations. All those parameters may be modelled as continuous variables and they may be integrated in different ways. Expenditure of effort and/or the generation of worthwhile effects may be the joint result of several of those parameters and they may heavily depend on the context.

### 3.4 *Location of Relevance*

In addition, notions of effort and effect have been used for different phases and actors in the translation process.

Table 2 indicates where translation effort and translation effect –and thus phenomena of relevance– have been suggested to occur: within the audience or/and within the translator. Gutt (1989; 2000), for instance, analyses textual features of *the translation product* as proxy for assumed processing effort and cognitive effects of a hypothetical *translation receiver*. In his framework, effort, effects and thus relevance are approximated qualitatively and mainly subjectively (through linguistic analysis), which makes this approach difficult to quantify or falsify.

| **Effort (measures)** | **Effect (measures)** |
|---|---|
| Receiver's Mind (properties of TT) | Receiver's mind (properties of TT) |
| Translator's behavior (keystrokes / gaze) | Receiver's mind (properties of TT) |
| Translator's behavior (keystrokes / gaze) | Target text (translation quality) |
| Translator's perception (properties of ST) | Translator's brain activity (brain imaging) |

Table 2: Location of effort and effects / relevance on the side of the receiver audience or translator

Indicators of effort have also been measured in the *translators' behavior* as captured by gaze movements (Alves and Vale 2009; Carl 2016) and/or keystroke pauses (e.g., Lacruz and Shreve 2014). This line of research opens the possibility to quantify effort and has given rise to empirical TPR and the TPR-DB, as outlined above. There are also various suggestions as to how translational *effects* should be measured. As Gonçalves (2020) points out, a troublesome assumption is the temporal and spatial dissociation of effort and effects which locates the effort in translators' behavior and the effects in the receivers' minds. This makes an assessment of relevance and the determination of optimum relevance practically impossible.

One approach to overcome this dissociation has been to measure translational effects as properties of the produced translation: either in the final translation product or in the typing itself. This allows for seamless (co)relation as effort (in terms of behavior) and effect parameters can be immediately available to the researcher. This has been a productive path of enquiry which has produced many studies and insights. A variation of this approach includes (retrospective) interviews to gauge effort and effect with translator's satisfaction self-assessment.

However, measuring translation effects by means of characteristics of the translation product presumes that a translation brief (or guidelines) is available (either implicitly or explicitly) by which translators are able to produce the desired translation quality that is believed to trigger the expected effects in an (assumed) receiver's mind. The assumption of a translation brief, thus, allows researchers to investigate aspects of effort and effects, observed during a translation session, independently from effort and effects for an audience, which would then fall under reception studies, see e.g., Whyatt (in print).

Some recent studies conceptualize translation effects in the *translator's brain* (mental actions), which can be measured using brain imaging technologies (e.g., fMRI, fNIRS, EEG). This approach assumes that the activation of different brain areas are effects of different processing mechanisms, triggered by targeted and reproducible stimuli. By gathering evidence from other brain studies, conclusions can be drawn about the properties and relations between these processes, and their relevance in terms of effort/effects. In some sense, this approach parallels the effort/effect assumptions in the receivers mind. However, it is unclear how all these different notions of relevance correlate.

## 4  Free Energy Principle and Active Inference

RT assumes that, due to the selection pressure towards increasing efficiency,

> the human cognitive system has developed in such a way that our perceptual mechanisms tend to automatically pick out relevant stimuli, our memory retrieval mechanisms tend automatically to activate potentially relevant assumptions, and our inferential mechanisms tend spontaneously to process them in the most productive way (Hedberg 2004)

Thus, RT is a "normative approach" to communication and translation, which appeals to an optimality criterion (Parr et al. 2023). However, RT rejects the possibility to measuring relevance and thus to quantify optimality. The FEP (Friston 2009; 2010) and AIF (Parr et al. 2022), in contrast, explain similar normative assumptions in a rigorous mathematical framework, providing a general theory as to why our cognitive systems must have evolved so as to automatically maximize relevance.

The Free Energy Principle (FEP) and Active Inference (AIF) are two closely related theoretical frameworks that have recently gained attention in cognitive science and neuroscience. These frameworks provide a unifying perspective on how biological organisms maintain their viability in the face of an uncertain and ever-changing environment. The FEP (Friston 2009; 2010) is a theory of how living systems maintain a balance between their internal states and the external environment. It posits that biological systems are constantly seeking to minimize the discrepancy between their expected and actual sensory input, which is captured by a quantity called *free energy*. This term is related to the amount of energy required to move the system from a baseline configuration into a (more) desirable configuration, which allows for smooth interaction with its environment. The reduction of free energy is said to be crucial for survival and for an agent to stay in its "phenotypical niche".

According to FEP (Parr et al. 2022, 24-26), free energy can be reduced in two different ways: either by updating our beliefs about the external world or by acting on the world to make it more similar to our preferences. Updating beliefs comes as a consequence of observations (i.e., through perception) as we reduce the difference between the prior beliefs (before receiving new input) and the posterior beliefs (after receiving new input). In the second case, we do not change our beliefs, but modify the world so as to generate successive observations that fit our expectations (or preferences). In this view, *cognitive effort* can be defined in terms of updating beliefs and the reduction in the gap between prior and posterior beliefs (Parr et al. 2023), while *effects* are the results of modifications in the world. Optimality (e.g., optimal relevance) can then be defined as a trade-off, i.e., a cost function, to optimize (or minimize) the relation between these two quantities, which is known as variational free energy.

The FEP has been applied to many areas of research, including perception, action, learning, and decision-making. AIF (Parr et al. 2022), on the other hand, is a computational framework that models how organisms can achieve optimum relevance by making predictions about the world and taking actions that minimize free energy. This framework has also been applied to a wide range of cognitive processes, including perception, attention, decision-making, and learning (Friston et al. 2017; Parr et al. 2022, chapter 10).

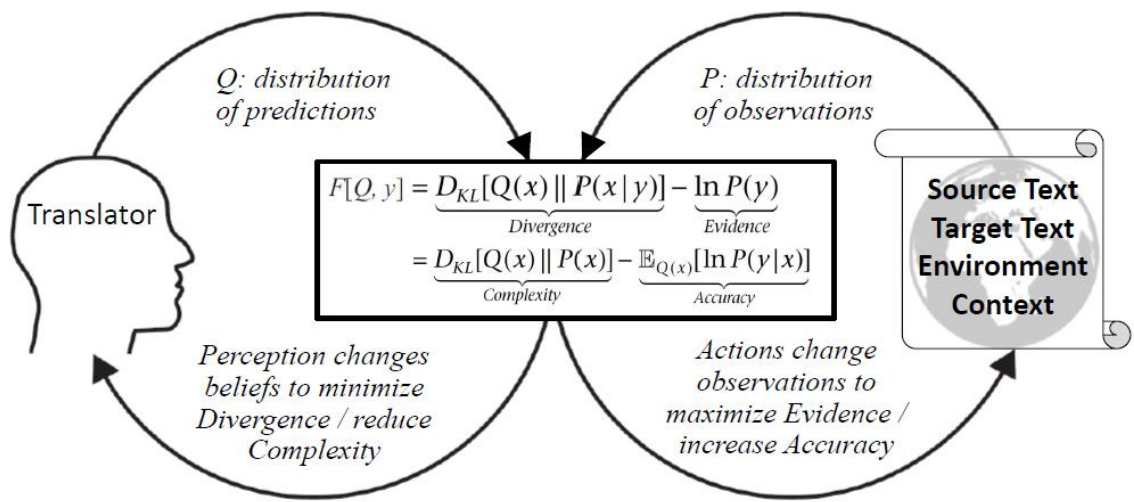

Figure 3: Free energy *F[Q,y]* in different factorizations, adapted from (Parr et al. 2022)

Figure 3 illustrates the FEP in a translation context. It shows the two factors that impact the discrepancy between the translator's expectations $Q$ of a (translation) task and the distribution $P$ of task related observations in an 'outside world'. $Q$ is a probability distribution over beliefs[6] ($x$) that takes into account the translator's habits (e.g., related to expertise), preferences (as communicated e.g., via a translation brief), skills (e.g., knowledge of translation tools, etc.), preferred strategies, etc.; it quantifies a *distribution* of the translator's beliefs concerning the task. The term $P(x \mid y)$ quantifies the probability of a belief $x$ when observing $y$. The term $P(y \mid x)$ quantifies the probability of an observation $y$ under the belief $x$. The Kullback-Leibler divergence ($D_{KL}$) quantifies the difference between the distributions $Q$ and $P$. If the translator has no information about the translation task whatsoever, $Q$ should have maximum entropy (i.e., all 'beliefs' are equally probable), which provides maximal compatibility with any and all possible observations and produces the smallest $D_{KL}$ value. The Kullback-Leibler divergence is minimal (zero) if the two distributions are identical, that is, if the observations fully coincide with the expectations.

---

[6] The term 'belief' is used here in a very wide sense, it includes, for instance, a translation relation, syntactic structure, textual continuation, translation guidelines, communication goal, etc.

The *Divergence* term in the first line of equation in Figure 3 scores the amount of belief updating (also called Bayesian surprise) when confronted with a new observation. In our definition above, this amounts to (translation) effort. This divergence is zero if the predictions fully coincide with the observations. In this case, free energy, ($F[Q, y]$) amounts to the *Evidence*, that is, the negative log probability - or Shannon surprise ($-\ln(P(y))$) - of the observation. The formulation of the free energy in the first line of the equation in Figure 3 thus suggests reducing Bayesian surprise by approximating $Q$ as much as possible.

Another, equivalent, formulation of free energy is shown in the second line in that figure. The *Complexity* term, just as the *Divergence*, quantifies Bayesian surprise, indicating how much a translator has to update their beliefs following an observation. The *Accuracy* term quantifies the conditional (Shannon) surprise of an observation weighted by predictions $Q(x)$. Crucially, the maximization of *Accuracy* depends on how much the outcomes of previous (translation) actions (that is, the effects) are in tune with forthcoming beliefs so as not to increase surprise (i.e., effort). That is, an action reduces entropy if it anticipates/coincides with an upcoming belief. Thus, the effect (of an action) can be assessed by the effort it generates during successive observation. Parr et al. (2023, 2) specify that the

> complexity quantifies the degree to which we must update our prior beliefs to explain the data at hand. This must be offset against the accuracy with which we can predict those data … [actions] allow us to modify the data we will receive in the future, and so decisions about which action to take must be based upon anticipated data.
> Parr et al. 2023, 2

In this sense, an agent will selectively sample the sensory inputs that it expects (Friston 2010), and it will aim at producing changes in the environment that are in tune with her predictions. In other terms, an individual self-evidences herself by acting on the world to implement her own preferences (Kirchhoff and Kiverstein 2021).

AIF stipulates that agents follow *action policies (π)* that specify (probably hierarchically organized) sequences of action. From a set of possible action policies, an agent will select the one that maximally reduces the *expected free energy* (Friston et al. 2017; Pezzulo et al. 2018). In this sense, AIF "assumes that both *perception and action cooperate to realize a single objective*—or optimize just one function—rather than having two distinct objectives" (Parr 2022, 24, emph. in original).

The hierarchical organization of the action policies allows us to conceive of "deep temporal" cognitive architectures in which different processes run in different timelines. Thus, in the translation context, the AIF framework makes it possible to formalize fast processes that realizes horizontal/s-mode translations following sets of preferences (e.g., priming mechanisms) and assumptions, which may be interrupted by slower i-mode processes that are driven by different preferences or habits. FEP and AIF provide, thus, a formal framework to Monitor Model.

## 5 Discussion

The paper reviews the development of research and modelling approaches in empirical TPR and elaborates a novel view on the translation process as a possible framework for future research. The paper points out compatibilities between the Monitor Model (Tirkkonen-Condit 2005; Schaeffer and Carl 2013/2015), Relevance Theory (Gutt 2000; 2004: 2005) and the FEP/AIF (Friston 2010; Parr et al. 2022), and their complementary with respect to (1) a temporal stratification of different processing layers that presumes different timelines for distinct processing routes, (2) a normative approach that assumes an optimization of (translation) behavior which is based in fundamental principles of cognition (or even life in general).

The Monitor Model draws on a body of research in psycholinguistic studies, suggesting that automated translation routines are associated with comparatively low levels of effort, as they involve minimal cognitive resources. The Monitor Model suggests that much of basic translation production emerges out of these *horizontal,* automated processes, but these horizontal translation routines may be interrupted and complemented/controlled by higher-level (monitoring) or *vertical* processes (Tirkkonen-Condit 2005). These distinct processes involve different cognitive resources and evolve on different timelines.

Drawing on Relevance Theory (RT), Gutt (2004; 2005) describes translation as a form of (cross-linguistic) *interpretive language use*. Translation is, according to him, an act of (re)producing cross-linguistic similarity, rather than describing or evaluating the truth of source language statements. Gutt suggests two modes of interpretive language use: the *stimulus mode* (s-mode, "what was said") which informs the target audience about the linguistic properties of the source vs. the *interpretive mode* (i-mode, "what was meant") which addresses the intended interpretation. S-mode translations rely on the similarity of the linguistic properties in the input and output while i-mode translations rely on the similarity of their interpretations. Gutt does not specify how these two modes interact during the translation process. In this paper, I suggest that the s-mode and i-mode roughly correspond to horizontal and vertical processes, respectively, as proposed in the Monitor Model. In this view, the s-mode serves as the basis of translational production which is interrupted by the i-mode to account for (better) interpretive resemblance. These concepts are related to *default* translations (Carl and Dragsted 2012; Halverson 2019) or the *cruise mode* (Pym 2017), a processing mode which Pym characterizes as "all goes well until there is a 'bump', attention is required, and something needs to be done" (cf. Tirkkonen-Condit 2005).

A central concept in RT is the notion of relevance. RT defines relevance as the trade-off between cognitive effort and effects. Lower cognitive effort combined with higher effects leads to increased relevance, while higher cognitive effort and/or lower effects decrease the relevance of a communicative act. RT stipulates that relevance is a principle, rather than a maxim, that is sought to be followed, but does not provide clear measures for quantifying relevance.

The FEP, in contrast, provides a framework that is suited to quantify relevance in terms of free energy. The FEP, and as its corollary AIF (Parr et al. 2022), constitute a framework to quantify the discrepancy between action and perception as surprise (or its upper bound, free energy). Following Parr et al. (2022) "surprise minimization can be construed as the reduction of the discrepancy between the model and the world." (39). Parr and colleagues (2023) show how this can be achieved in 'deep temporal' systems that assume concurrent processes in multiple timelines. In future, we will discuss in detail how deep temporal translation processes can be modelled within FEP/AIF.

According to FEP, an agent aims at reducing free energy by adjusting its internal model in line with the perceived observations, and/or by changing the environment so that it becomes more similar to her preferences. Observations that are in tune with expectations and actions that are in tune with preferences (or habits) impliy least cognitive effort. Cognitive effort can then be conceptualized as "the qualitative experience of committing to a behavior that diverges from priori habits" (Parr et al. 2023). Parr et al model cognitive effort as the divergence between context sensitive beliefs about how to act and a context insensitive prior belief such as habits or preferences. Behavioral patterns that correspond to habits (or preferences) can thus be assumed to exert least effort, while behavioral patterns that involve - or result from - changes in the agent's belief system are expected to be more effortful.

In this view, the trade-off between the cognitive effort and effects (and thus relevance) can be measured as the best fit between the predicted and observed sensory input, and is thus a special case of free energy. Provided the s-mode and the i-mode result in comparable translational effects for a hypothetical audience, the s-mode turns out to be the more relevant one. In other words, assuming similar effects, s-mode translations can be expected to be less effortful as they do not involve an evaluation or change of the translator's belief system, as the i-mode translations do.

There are limits and interesting predictions to this approach that may be worth adressing in future TPR. Discussing difficulties of Bible translation for speakers of an Ethiopian language, Gutt (2005) maintains that:

> either the audience's cognitive environment [i.e., their prior beliefs] needs to be adjusted so that it can process this information, or this information needs to be left aside in the higher-order communication act.
> Gutt 2005: 47

That is, either the audience's model (i.e., their cognitive environment) has to be adjusted or the translation for those speakers has to change (e.g., left aside), since "it is entirely unreasonable to expect that information for which their cognitive environment is not prepared can be communicated" (Gutt, 2005). In the context of AIF, Parr et al. (2023, 3) come to very similar conclusions when they find that "cognitive effort must be deployed to overcome a habit that is incongruent with our goals, but a sufficiently strong habit prevents deployment of effort to overcome that habit." Future predictive TPR should be able to assess such situations.